\documentclass[conference]{IEEEtran}
\IEEEoverridecommandlockouts

\usepackage{cite}
\usepackage{amsmath,amssymb,amsfonts}
\usepackage{algorithmic}
\usepackage{graphicx}
\usepackage{textcomp}
\usepackage{xcolor}
\def\BibTeX{{\rm B\kern-.05em{\sc i\kern-.025em b}\kern-.08em
    T\kern-.1667em\lower.7ex\hbox{E}\kern-.125emX}}
\begin{document}

\title{Enhancing Automated Paper Reproduction via Prompt-Free Collaborative Agents\\
\thanks{$^\dagger$These authors contributed equally.}
\thanks{$^*$Corresponding author.}
}

\author{
Zijie Lin$^{1, 2,\dagger}$, Qilin Cai$^{1,\dagger}$, Liang Shen$^{2}$, Mingjun Xiao$^{1,*}$\\
\textit{$^1$University of Science and Technology of China, Hefei, China}\\
\textit{$^2$Meituan, Beijing, China}\\
Emails: \{lzj741, cql111\}@mail.ustc.edu.cn, shenliang03@meituan.com, xiaomj@ustc.edu.cn\\
}

\maketitle

\begin{abstract}
Automated paper reproduction has emerged as a promising approach to accelerate scientific research, employing multi-step workflow frameworks to systematically convert academic papers into executable code. However, existing frameworks often lack mechanisms to verify and refine the outputs at each generation step, or rely heavily on manually designed prompts for self-refinement, which limits their adaptability and scalability. To address these limitations, we propose a prompt-free collaborative agent framework that automatically enhances the quality of paper-to-code generation. Our approach employs two collaborative agents: a verification agent that examines whether the outputs at each step satisfy the requirements specified in the corresponding system prompt, and a refinement agent that revises the outputs based on the identified issues. Unlike previous methods that require human experts to craft specific refinement prompts for each step, our framework achieves automatic verification and improvement by leveraging only the original system prompts. We integrate our collaborative agents into the Paper2Code framework and conduct comprehensive experiments on PaperBench Code-Dev and Paper2CodeBench datasets. Experimental results demonstrate that our approach significantly improves the accuracy and completeness of reproduced code, achieving performance gains of approximately 15\% and 13\%, respectively, compared to the baseline without our agents. Furthermore, comparative experiments against Self-Refine validate the robustness and consistency of our prompt-free approach across different datasets.
\end{abstract}

\begin{IEEEkeywords}
Collaborative Agents, Paper Reproduction, Prompt-free Refinement, Code Generation.
\end{IEEEkeywords}

\begin{figure*}[t] 
  \centering       
  \includegraphics[width=0.9\textwidth]{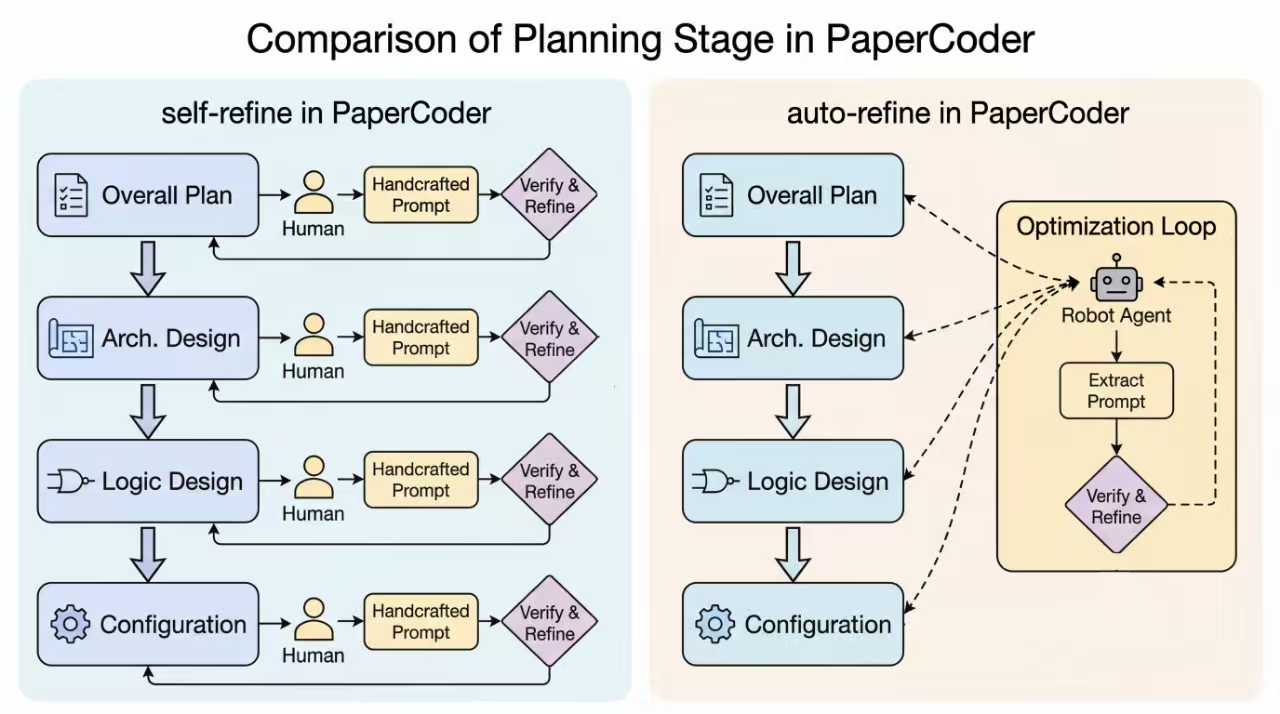} 
  \caption{Comparison of the planning stage in PaperCoder between the self-refine and auto-refine methods.
The left panel illustrates the self-refine approach, where each planning step requires human-crafted prompts for verification and refinement. The right panel presents our proposed auto-refine method, which employs an automated optimization loop. In this approach, a unified agent automatically extracts prompts to verify and refine the plan at each stage, eliminating the need for manual intervention.}
  \label{fig1}
\end{figure*}

\section{Introduction}
The reproducibility of scientific research has become a critical concern in the modern research landscape, particularly as the volume and complexity of published work continue to grow exponentially \cite{b1,b2}. Reproducing experimental results from academic papers requires substantial time and expertise, often taking weeks or months for researchers to implement and validate the methods described in publications \cite{b4}. The lack of publicly available code implementations further impedes the reproduction process \cite{b18}. This is particularly evident in rapidly advancing fields: only an average of 19.5\% of papers accepted to top-tier machine learning conferences in 2024 provide their code implementations \cite{b3}. This reproducibility gap not only slows scientific progress but also raises questions about the reliability and verifiability of research findings across disciplines \cite{b15,b16}.

Recent advances in large language models have opened new possibilities for automating the paper reproduction process \cite{b11,b12,b13}. These models demonstrate remarkable capabilities in understanding natural language descriptions and generating corresponding code implementations \cite{b14,b17,b30}. Consequently, automated paper reproduction systems have emerged as a promising solution to bridge the gap between theoretical descriptions in academic papers and their practical implementations \cite{b21,b22,b33}. By systematically converting research papers into executable code, these systems have the potential to accelerate scientific discovery, facilitate knowledge transfer, and enhance research reproducibility across multiple domains \cite{b23,b24,b25}.

Current automated paper reproduction frameworks typically employ multi-step workflow architectures \cite{b7,b8,b26}, where the reproduction process is decomposed into sequential stages: planning, which determines the structure and purpose of code files; analyzing, which specifies detailed design specifications for each file; and coding, which implements the actual code based on the established specifications \cite{b19,b20}. Each stage receives specific instructions through system prompts that guide the generation process. While this structured approach has shown promising results, existing frameworks face two fundamental limitations that hinder their practical deployment and effectiveness. First, most frameworks lack systematic mechanisms to verify and refine outputs at each generation step \cite{b7}. Without proper validation, errors introduced in early stages can propagate through the workflow, leading to incorrect or incomplete final implementations \cite{b28,b29}. Although some recent work has explored self-refinement techniques to address this issue, these approaches rely heavily on manually designed prompts that specify how to check and improve outputs at each step \cite{b9,b10,b27}. This manual prompt engineering process requires substantial human expertise and effort, limiting the scalability and adaptability of these systems. Moreover, the quality of refinement heavily depends on the expertise of prompt designers, making it difficult to maintain consistent performance across different tasks and domains. Second, the dependence on human-crafted refinement prompts creates a bottleneck in the automation pipeline \cite{b31,b32}. As paper reproduction workflows become more complex with additional stages and diverse task requirements, designing and maintaining appropriate refinement prompts for each step becomes increasingly challenging. This limitation is particularly problematic in collaborative research environments where multiple researchers with varying levels of expertise need to utilize these systems efficiently.

To address these challenges, we propose a prompt-free collaborative agent framework that automatically enhances the quality of paper-to-code generation without requiring manual prompt engineering for refinement. Our approach introduces two collaborative agents that work in tandem: a verification agent that examines whether the outputs at each workflow step satisfy the requirements specified in the corresponding system prompt, and a refinement agent that revises the outputs based on identified issues. The key insight underlying our approach is that system prompts already contain comprehensive specifications of what the outputs should include and how they should be structured. Rather than designing separate evaluation criteria and refinement instructions, we can directly use these existing requirements as verification benchmarks and improvement guidelines, ensuring perfect alignment between generation objectives, validation standards, and refinement goals. We integrate our collaborative agent framework into the Paper2Code system and evaluate its effectiveness on two benchmark datasets: PaperBench Code-Dev and Paper2CodeBench. Our experimental results demonstrate significant improvements in both accuracy and completeness of reproduced code, with performance gains of approximately 15\% and 13\%, respectively, compared to baselines without our agents. Furthermore, comparative experiments against existing self-refinement approaches validate the robustness and consistency of our prompt-free method across different datasets and task scenarios.

The main contributions of this work are threefold: (1) We identify the critical limitations of existing paper reproduction frameworks regarding output verification and prompt-dependent refinement. (2) We propose a novel prompt-free collaborative agent framework that achieves automatic verification and refinement by leveraging only the original system prompts. (3) To the best of our knowledge, this is the first work that explores prompt-free collaborative agents for refining multi-agent outputs in automated paper reproduction. Through comprehensive experiments, we demonstrate that our approach significantly improves reproduction quality. Our framework has the potential to provide a new optimization paradigm for automated code generation systems and inspire future research in prompt-free multi-agent collaboration.

\section{Method}
Figure~\ref{fig1} shows the difference between our framework and self-refine. Our framework consists of two collaborative agents: a verification agent that identifies issues by checking outputs against system prompt requirements, and a refinement agent that automatically improves outputs based on the identified issues. We describe each component in detail below.

\subsection{Verification Agent: Prompt-Free Output Validation}

The verification agent automatically validates generation outputs without requiring manually designed evaluation prompts. Let $\mathcal{P}$ denote the research paper content and $\mathcal{S}_i$ denote the system prompt for the $i$-th generation step in the paper reproduction workflow. At step $i$, the model produces an output $\mathcal{O}_i$ based on inputs $\{\mathcal{P}, \mathcal{S}_i, \mathcal{O}_{<i}\}$, where $\mathcal{O}_{<i}$ represents outputs from previous steps. Our verification agent $\mathcal{V}$ examines whether $\mathcal{O}_i$ satisfies the requirements specified in $\mathcal{S}_i$. Formally, the verification process can be expressed as:
$$
\mathcal{R}_i = \mathcal{V}(\mathcal{O}_i, \mathcal{S}_i, \mathcal{P})
$$
where $\mathcal{R}_i$ is a structured review report containing identified issues and improvement suggestions. In contrast to Self-Refine, which requires human experts to manually design evaluation rubrics for each generation step, our verification agent directly uses the original system prompt $\mathcal{S}_i$ as the evaluation standard. The verification agent systematically checks whether each requirement specified in $\mathcal{S}_i$ is met in $\mathcal{O}_i$. This design offers a critical advantage: guaranteed alignment between generation objectives and verification standards, as both processes reference the same system prompt $\mathcal{S}_i$.

The verification prompt is constructed by combining three key components: (1) \textbf{Paper Content} ($\mathcal{P}$), (2) \textbf{System Prompt Requirements} ($\mathcal{S}_i$), and (3) \textbf{Current Output} ($\mathcal{O}_i$). The template can be formalized as $\text{Prompt}_{\mathcal{V}} = f_{\text{template}}(\mathcal{P}, \mathcal{S}_i, \mathcal{O}_i)$. The verification agent produces a structured JSON report $\mathcal{R}_i = \{\text{completeness\_summary}, \text{missing\_information}, \text{action\_items}\}$, where \textbf{missing\_information} is a list $[m_1, m_2, ..., m_k]$ of specific issues where each $m_j$ describes a requirement from $\mathcal{S}_i$ that is not adequately satisfied in $\mathcal{O}_i$, and \textbf{action\_items} is a list $[a_1, a_2, ..., a_k]$ of concrete improvement suggestions where each $a_j$ corresponds to addressing issue $m_j$. This structured format enables the subsequent refinement agent to systematically address identified problems.

\subsection{Refinement Agent: Automatic Output Improvement}

The refinement agent automatically improves generation outputs based on the structured review reports produced by the verification agent. Let $\mathcal{R}_i = \{\text{completeness\_summary}, \\\text{missing\_information}, \text{action\_items}\}$ denote the verification report for output $\mathcal{O}_i$. The refinement agent $\mathcal{F}$ takes the original output $\mathcal{O}_i$, the verification report $\mathcal{R}_i$, and the original system prompt $\mathcal{S}_i$ as inputs to produce an improved output $\mathcal{O}_i^*$. Formally, the refinement process can be expressed as:

$$
\mathcal{O}_i^* = \mathcal{F}(\mathcal{O}_i, \mathcal{R}_i, \mathcal{S}_i, \mathcal{P})
$$

In contrast to Self-Refine, which requires human experts to manually design refinement instructions for each generation step, such as specifying how to improve completeness, how to enhance clarity, and how to better align with requirements, our refinement agent directly uses the original system prompt $\mathcal{S}_i$ as the refinement guidance. The key insight is that $\mathcal{S}_i$ already contains all the specifications that the output should satisfy, and the verification report $\mathcal{R}_i$ identifies exactly which specifications are not met. Therefore, the refinement agent can systematically address each issue in $\mathcal{R}_i$ by ensuring the refined output $\mathcal{O}_i^*$ satisfies the corresponding requirements in $\mathcal{S}_i$.

The refinement prompt is constructed by combining five key components: (1) \textbf{Paper Content} ($\mathcal{P}$), (2) \textbf{System Prompt Requirements} ($\mathcal{S}_i$), (3) \textbf{Original Output} ($\mathcal{O}_i$), (4) \textbf{Verification Report} ($\mathcal{R}_i$), and (5) \textbf{Previously Refined Outputs} ($\{\mathcal{O}_1^*, \mathcal{O}_2^*, ..., \mathcal{O}_{i-1}^*\}$) for maintaining consistency across multi-step workflows. The template can be formalized as $\text{Prompt}_{\mathcal{F}} = f_{\text{refine}}(\mathcal{P}, \mathcal{S}_i, \mathcal{O}_i, \mathcal{R}_i, \{\mathcal{O}_j^*\}_{j<i})$. Specifically, the prompt instructs the refinement agent to: (1) preserve the correct parts of $\mathcal{O}_i$ that already satisfy $\mathcal{S}_i$, (2) address each issue listed in $\mathcal{R}_i$.\textbf{missing\_information} by incorporating the missing requirements from $\mathcal{S}_i$, (3) follow the improvement suggestions in $\mathcal{R}_i$.\textbf{action\_items} while ensuring compliance with $\mathcal{S}_i$, and (4) maintain consistency with previously refined outputs. This design ensures that refinement is guided by the same requirements used for generation and verification, guaranteeing alignment across the entire workflow.


\begin{table*}[htbp]
\caption{Summary Statistics of Optimization Performance on PaperBench (with LLM details). \textsuperscript{*}Results for RePro are cited from the original paper as the code is not open-source.}
\begin{center}
\begin{tabular}{|l|c|c|c|c|c|c|}
\hline
\textbf{Method} & \textbf{LLM} & \multicolumn{2}{c|}{\textbf{Absolute Scores}} & \multicolumn{3}{c|}{\textbf{Comparison vs. Paper2code}} \\
\cline{3-7}
& & \textbf{\textit{Average}} & \textbf{\textit{Median}} & \textbf{\textit{Win Rate}} & \textbf{\textit{Avg. Improvement}} & \textbf{\textit{Max Improvement}} \\
\hline
RePro\textsuperscript{*} & o3-mini-high & 0.626 & - & - & - & - \\
\hline
paper2code & GPT-4.1 & 0.682 & 0.692 & - & - & - \\
\hline
paper2code + self-refine in plan & GPT-4.1 & 0.655 & 0.655 & 10/20 (50.0\%) & -3.96\% & +32.08\% \\
\hline
paper2code + auto-plan optimized & GPT-4.1 & 0.723 & 0.768 & 11/20 (55.0\%) & +6.01\% & \textbf{+58.23\%} \\
paper2code + auto-code optimized & GPT-4.1 & 0.747 & 0.787 & 16/20 (80.0\%) & +9.53\% & +42.98\% \\
paper2code + auto-plan \& code optimized & GPT-4.1 & \textbf{0.786} & \textbf{0.827} & \textbf{17/20 (85.0\%)} & \textbf{+15.25\%} & +56.88\% \\
\hline
\end{tabular}
\label{tab:paperbench_stats}
\end{center}
\end{table*}

\begin{table*}[htbp]
\caption{Performance Comparison of Different Optimization Strategies on the Paper2CodeBench Dataset}
\begin{center}
\begin{tabular}{|l|c|c|c|c|c|}
\hline
\textbf{Method} & \textbf{LLM} & \multicolumn{3}{c|}{\textbf{Conference Scores}} & \textbf{Statistics} \\
\cline{3-5} 
& & \textbf{\textit{ICLR 2024}} & \textbf{\textit{ICML 2024}} & \textbf{\textit{NeurIPS 2024}} & \textbf{\textit{Average}} \\
\hline
paper2code & GPT-4.1 & 3.85 & 4.09 & 3.60 & 3.84 \\
\hline
paper2code + self-refine in plan & GPT-4.1 & 4.14 & 4.26 & 3.75 & 4.05 \\
\hline
paper2code + auto-plan optimized & GPT-4.1 & 4.01 & 4.00 & 3.80 & 3.94 \\
paper2code + auto-code optimized & GPT-4.1 & 4.28 & 4.39 & 4.18 & 4.29 \\
paper2code + auto-plan \& code optimized & GPT-4.1 & \textbf{4.35} & \textbf{4.43} & \textbf{4.23} & \textbf{4.34} \\
\hline
\end{tabular}
\label{tab:optimization_results}
\end{center}
\end{table*}


\begin{table}[htbp]
\caption{Comparison of Performance and Efficiency against RePro. \textsuperscript{*}Results for RePro are cited from the original paper. P2C denotes paper2code.}
\begin{center}
\resizebox{\linewidth}{!}{
    \begin{tabular}{|p{3cm}|c|c|c|c|c|}
    \hline
    \textbf{Method} & \textbf{LLM} & \textbf{Iter.} & \textbf{Ori. Score} & \textbf{Final Score} & \textbf{Avg. Improve.} \\
    \hline
    RePro\textsuperscript{*} & o3-mini-high & 5 & 0.528 & 0.614 & +16.29\% \\
    \hline
    P2C + auto-plan \& code optimized & GPT-4.1 & 1 & 0.682 & 0.786 & +15.25\% \\
    \hline
    \end{tabular}
}
\label{tab:paperbench_stats_simplified}
\end{center}
\end{table}

\subsection{Application to Paper2Code Workflow}

We apply our collaborative agent framework to two critical stages in the Paper2Code workflow: the planning stage and the coding stage. The planning stage produces four key artifacts: (1) \textbf{Overall Plan} ($\mathcal{O}_1$), which outlines the methodology and experimental setup; (2) \textbf{Architecture Design} ($\mathcal{O}_2$), which specifies the software system structure and file organization; (3) \textbf{Logic Design} ($\mathcal{O}_3$), which decomposes tasks and analyzes dependencies; and (4) \textbf{Configuration} ($\mathcal{O}_4$), which captures hyperparameters and training settings. The coding stage generates Python implementation files and a configuration file (config.yaml) based on the planning artifacts. For each artifact, the corresponding system prompt from Paper2Code's original workflow serves as both the verification standard and refinement guidance.

The verification-refinement process operates sequentially to maintain consistency across artifacts. For the planning stage, we first verify the Overall Plan using $\mathcal{V}(\mathcal{O}_1, \mathcal{S}_1, \mathcal{P})$ to produce review report $\mathcal{R}_1$, then refine it using $\mathcal{F}(\mathcal{O}_1, \mathcal{R}_1, \mathcal{S}_1, \mathcal{P})$ to produce $\mathcal{O}_1^*$. This refined output is then used as input for subsequent artifacts: the Architecture Design is verified and refined based on $\mathcal{O}_1^*$, producing $\mathcal{O}_2^*$; the Logic Design is verified and refined based on both $\mathcal{O}_1^*$ and $\mathcal{O}_2^*$, producing $\mathcal{O}_3^*$; and finally, the Configuration is verified and refined based on all previous refined artifacts, producing $\mathcal{O}_4^*$. This sequential process ensures that later artifacts can reference and maintain consistency with earlier refined artifacts.

For the coding stage, the verification-refinement process follows the task dependency order specified in the Logic Design ($\mathcal{O}_3^*$). Each Python implementation file is verified against the coding system prompt requirements, which include writing elegant and modular code, strictly adhering to the paper's methodology, implementing complete code without TODO placeholders, following the design specifications from the planning stage, and correctly referencing configurations from config.yaml. The refinement agent then improves each code file by incorporating previously refined code files into the refinement context, ensuring that the entire codebase forms a coherent implementation. The refined outputs are written back to their original locations, enabling subsequent stages to immediately benefit from the refinements. While Self-Refine requires manually designing separate evaluation and refinement prompts for each artifact—resulting in 8 hand-crafted prompts for the 4 planning artifacts alone in the original Paper2Code implementation—our approach automatically adapts to each artifact by directly leveraging its corresponding system prompt, requiring zero additional prompt engineering.

\begin{figure*}[t] 
  \centering       
  \includegraphics[width=0.9\textwidth]{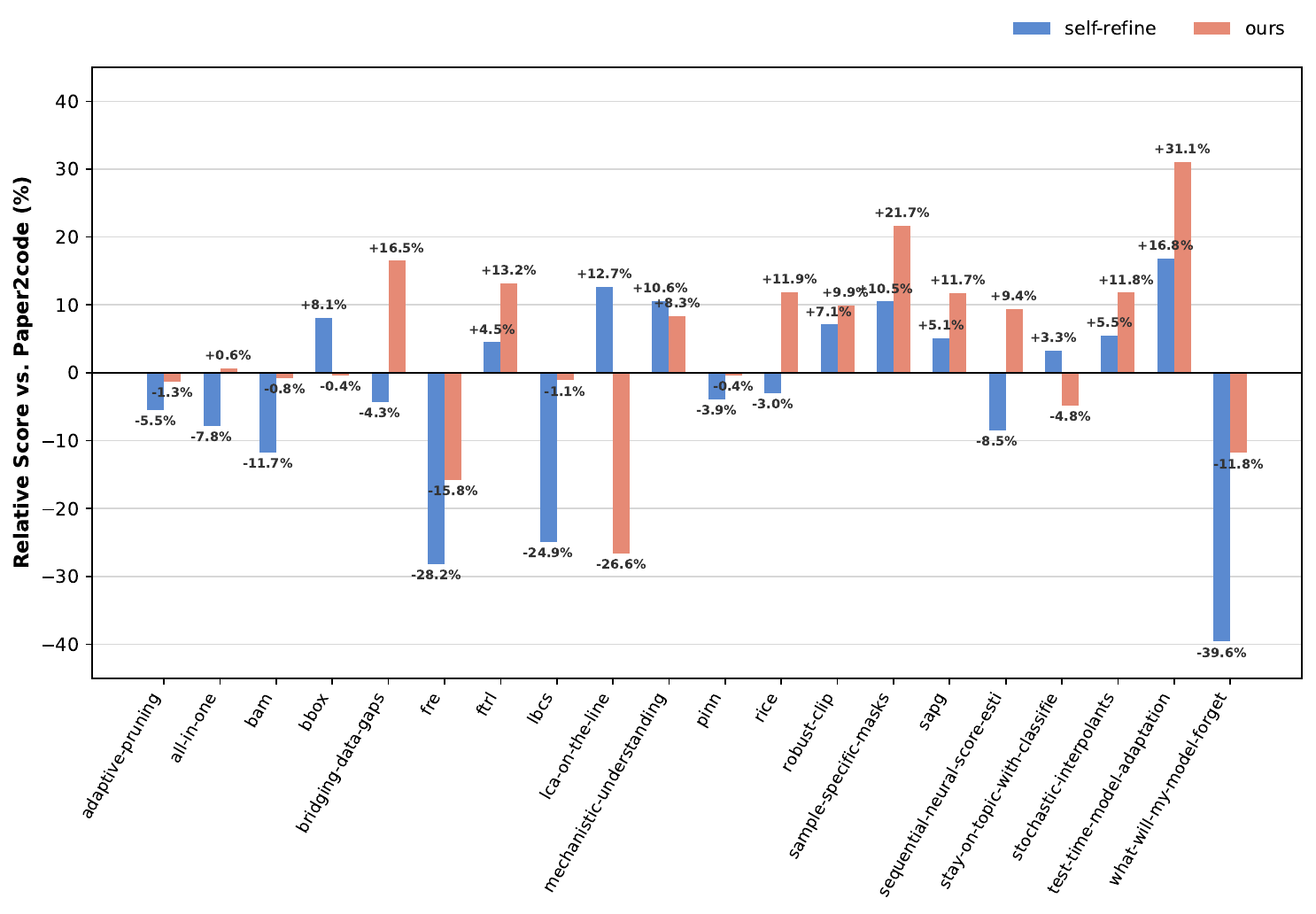} 
  \caption{Per-paper performance comparison on PaperBench between our planning optimization approach (orange) and Self-Refine (blue). Relative scores are calculated against the Paper2Code baseline.}
  \label{fig}
\end{figure*}

\section{Experiment}
\subsection{Datasets}
We conduct our evaluation on two benchmarks: \textbf{Paper2CodeBench} \cite{b3}, containing \textbf{90} papers from top-tier 2024 conferences (ICLR, ICML, NeurIPS) filtered by code availability and length, where our generated repositories are evaluated against official author-released implementations using LLM-based judges that assess correctness on a 5-point scale, and \textbf{PaperBench Code-Dev} \cite{b4}, containing \textbf{20} ICML 2024 papers with human-annotated grading rubrics for LLM-based functional correctness verification.

\subsection{Baselines}
To the best of our knowledge, there are few existing methods that employ iterative self-refinement specifically for automated paper-to-code reproduction. Therefore, we compare our approach against the two most relevant baselines that employ iterative refinement for paper-to-code reproduction:
(1) Paper2Code + Self-Refine \cite{b3}. Self-Refine \cite{b5} is an iterative refinement framework where LLMs critique and improve their own outputs. To ensure a fair comparison, we adopt the Self-Refine implementation from the original Paper2Code work, where the authors manually designed evaluation and refinement prompts for each of the four planning artifacts (Overall Plan, Architecture Design, Logic Design, and Configuration)—totaling 8 hand-crafted prompts. The evaluation prompts instruct the model to assess artifacts using structured rubrics (e.g., "Extract Paper Requirements → Map Requirements to Plan → Assess Success Criteria → Critique → Score: Provide a 1–5 rating"), while refinement prompts guide improvements based on the critiques. In contrast, our approach requires zero additional prompt engineering by directly leveraging the original system prompts.
(2) RePro \cite{b6}. RePro automatically extracts a paper's "fingerprint"—a comprehensive set of accurate and atomic criteria—as supervisory signals. It first generates code based on extracted information, then uses the fingerprint within an iterative verification-refinement loop to detect discrepancies and produce targeted revisions. Similar to our approach, RePro employs iterative refinement to improve code quality.

\subsection{Experiment Settings}
We conduct all experiments using GPT-4.1 as the foundation model. For the RePro baseline, due to the unavailability of its source code, we directly report the results from the original paper, which were obtained using o3-mini-high.




\subsection{Main Results}

Table~\ref{tab:paperbench_stats} and Table~\ref{tab:optimization_results} present the performance of different optimization strategies on PaperBench Code-Dev and Paper2CodeBench, respectively. \textbf{First, our approach demonstrates superior robustness across datasets.} When optimizing only the planning stage, our method (paper2code + auto-plan optimized) achieves consistent improvements on both benchmarks: +6.01\% average improvement on PaperBench (Table~\ref{tab:paperbench_stats}) and an average score of 3.94 on Paper2CodeBench (Table~\ref{tab:optimization_results}). In contrast, Self-Refine exhibits inconsistent performance across datasets. While Self-Refine achieves larger improvements on Paper2CodeBench (+0.21 average score improvement from 3.84 to 4.05), it actually degrades performance on PaperBench (-3.96\% average improvement with only 50\% win rate). This discrepancy can be attributed to the fact that the Self-Refine prompts in the original Paper2Code work were manually designed and tuned specifically for Paper2CodeBench, leading to overfitting to that particular dataset. When evaluated on the different distribution of PaperBench, these hand-crafted prompts fail to generalize, resulting in negative improvements. Our approach, by directly leveraging the original system prompts without additional prompt engineering, demonstrates better generalization and robustness across different paper distributions.

\textbf{Per-task analysis reveals consistent superiority.} Figure~\ref{fig} presents a detailed per-task breakdown of performance on PaperBench. Our planning optimization outperforms Self-Refine on 16 out of 20 tasks (80\%), demonstrating robust improvements across diverse paper reproduction scenarios. Notably, Self-Refine shows negative improvements on 10 tasks, with particularly severe degradations on tasks like \textit{what-will-my-model-forget} (-39.6\%) and \textit{fre} (-28.2\%), while our method exhibits more stable performance with smaller variance. 

For the 4 tasks where our planning optimization shows performance drops (e.g., \textit{mechanistic-understanding}: -26.6\%, \textit{fre}: -15.8\%), we observe a systematic pattern: our verification prompts prioritize overall paper completeness and high-level design correctness over implementation details such as hyperparameter settings and model loading procedures. This design choice, while beneficial for architectural coherence, can lead to lower scores on tasks where fine-grained implementation details are critical for evaluation. However, this limitation is effectively addressed by our coding-stage optimization. As shown in Table~\ref{tab:paperbench_stats}, combining both planning and coding optimizations (paper2code + auto-plan \& code optimized) achieves the best overall performance of 0.786, representing a 20\% relative improvement over Self-Refine's 0.655. This demonstrates that our two-stage optimization strategy successfully balances high-level design quality with implementation correctness.

\textbf{Second, our method exhibits strong extensibility across workflow stages.} Beyond planning optimization, we apply our verification-refinement framework to the coding stage (paper2code + auto-code optimized), achieving substantial improvements: +9.53\% on PaperBench with an 80\% win rate (Table~\ref{tab:paperbench_stats}) and an average score of 4.29 on Paper2CodeBench (Table~\ref{tab:optimization_results}). When combining both planning and coding optimizations (paper2code + auto-plan \& code optimized), we achieve the best overall performance: 0.786 average score on PaperBench (85\% win rate, +15.25\% improvement) and 4.34 average score on Paper2CodeBench. These results demonstrate that our approach can be seamlessly applied to different stages of multi-agent workflows without requiring stage-specific prompt engineering, highlighting its strong extensibility and practical applicability.

\textbf{Comparison with RePro.} As shown in Table~\ref{tab:paperbench_stats_simplified}, our method demonstrates superior efficiency: achieving +15.25\% improvement in a single iteration versus RePro's +16.29\% over 5 iterations—a 5× reduction in computational cost for comparable gains. Furthermore, our final score (0.786 with GPT-4.1) substantially outperforms RePro's (0.614 with o3-mini-high). This efficiency advantage is crucial for practical deployment, as each iteration involves costly LLM calls.

\section{Conclusion}
We proposed a prompt-free collaborative agent framework that automatically enhances automated paper reproduction quality without requiring manual refinement prompt design. Our approach employs verification and refinement agents that leverage only original system prompts, achieving significant improvements of approximately 15\% and 13\% on PaperBench Code-Dev and Paper2CodeBench, respectively. Comparative experiments demonstrated superior robustness over Self-Refine across different datasets. This work represents an initial exploration of prompt-free collaborative agents for multi-agent output refinement in paper reproduction, which may offer useful insights for automated code generation systems.

\end{document}